\documentclass[runningheads]{llncs}
\pdfoutput=1
\usepackage{graphicx}
\usepackage{booktabs}
\usepackage{amsmath}
\usepackage{multirow}
\usepackage{siunitx}
\usepackage{cite}

\begin{document}
\title{fMRI-S4: learning short- and long-range dynamic fMRI dependencies using 1D Convolutions and State Space Models}

\author{Ahmed El-Gazzar\inst{1,2} \and
Rajat Mani Thomas\inst{1,2} \and
Guido van Wingen\inst{1,2}}
\authorrunning{El-Gazzar et al.}
\titlerunning{fMRI-S4}
%
\institute{{Amsterdam UMC location University of Amsterdam, Department of Psychiatry} \and Amsterdam Neuroscience, Amsterdam, The Netherlands} 
\maketitle              
\begin{abstract}
Single-subject mapping of resting-state brain functional activity to non-imaging phenotypes is a major goal of neuroimaging. The large majority of learning approaches applied today rely either on static representations or on short-term temporal correlations. This is at odds with the nature of brain activity which is dynamic and exhibit both short- and long-range dependencies. Further, new sophisticated deep learning approaches have been developed and validated on single tasks/datasets. The application of these models for the study of a different targets typically require exhaustive hyperparameter search, model engineering and trial and error to obtain competitive results with simpler linear models. This in turn limit their adoption and hinder fair benchmarking in a rapidly developing area of research.   
To this end, we propose \textbf{fMRI-S4}; a versatile deep learning model for the classification of phenotypes and psychiatric disorders from the timecourses of resting-state functional magnetic resonance imaging scans. fMRI-S4 capture short- and long-range temporal dependencies in the signal using 1D convolutions and the recently introduced state-space models \textbf{S4}. The proposed architecture is lightweight, sample-efficient and robust across tasks/datasets. We validate fMRI-S4 on the tasks of diagnosing major depressive disorder (MDD), autism spectrum disorder (ASD) and sex classifcation on three multi-site rs-fMRI datasets. We show that fMRI-S4 can outperform existing methods on all three tasks and can be trained as a \textit{plug}\&\textit{play} model without special hyperpararameter tuning for each setting\footnote{Code available at https://github.com/elgazzarr/fMRI-S4}. 

\keywords{Functional Connectivity  \and State space models  \and  1D CNNs \and Major depressive disorder \and Autism spectrum disorder \and}
\end{abstract}
\section{Introduction}
Predicting non-imaging phenotypes from functional brain activity is one of the major objectives of the neuroimaging and neuroscience community. The ability to map a scan of the brain to behaviour or phenotypes would advance our understating of the brain function and enable the investigation of the underlying pathophysiology of psychiatric disorders. To build such prediction models, researchers have opted for machine learning to capture multivariate patterns in brain functional activity that might act as a bio-marker for the phenotype in question \cite{arbabshirani2017single, sundermann2014multivariate}. Functional magnetic resonance imaging (fMRI) offers a promising non-invasive tool to estimate brain functional activity by measuring the  blood-oxygenation-level-dependent (BOLD) signal as a proxy of the underlying neuronal activity of the brain \cite{ogawa1990brain}. From a machine learning point of view, fMRI is one of the most challenging data representations. The high dimensionality of the data (4D,  $\approx$ 1M voxels), low signal to noise ratio, data heterogeneity and limited sample sizes present major hurdles when developing learning models from fMRI data. To overcome some of these limitations, researchers have opted for summarized representations that facilitate learning from such data and can be interpreted for biomarker discovery.  One of the most popular representations is functional connectivity (FC) \cite{friston1994functional}. FC is simply defined as the temporal correlation in the BOLD signal changes between different regions of interest in the brain (ROIs). The study of FC have provided a wealth of knowledge about the brain function and dysfunction and have proved that there exists underlying neural correlates of phenotypes that a machine learning model can learn\cite{friston2011functional}. Yet, FC in its most popular form is a static representation which implicitly assumes that brain connectivity is static for the duration of the scan. In recent years, there has been growing evidence that brain connectivity is dynamic as the brain switch from cognitive states even at rest and more research is currently studying dynamic functional connectivity\cite{hutchison2013dynamic,preti2017dynamic}. With the advancement of deep learning, researchers have opted for sequential models such as RNNs, LSTMs, 1D CNNs and Transformers to learn from ROIs timecourses directly \cite{el2019simple,gadgil2020spatio, malkiel2021pre, yan2019discriminating, dvornek2017identifying, el2019hybrid}. Because the number of parameters of these models scale with the length of the timecourse and fMRI datasets are typically limited in sample sizes, applying such models to cover the entire duration of the resting-state scan is highly prone to overfitting. Further, recurrent models are not parallelizable and similar to attention-based models are computationally expensive to train. A practical solution is then to either i) crop the ROI timecourses, train the models on the cropped sequences, then aggregate the predictions from all the crops (e.g. via voting) to generate the final prediction at inference \cite{dvornek2017identifying, gadgil2020spatio, el2019hybrid}. or ii) limit the effective receptive field of the model using smaller kernels and fewer layers \cite{el2019simple}. The former method is prevalent with recurrent and attention based models while the latter is typically observed with convolutional models. fMRI BOLD signals display rich temporal organization, including scale-free $1/f$ power spectra and long-range temporal auto-correlations, with activity at any given time being influenced by the previous history of the system up to several minutes into the past \cite{he2011scale, hutchison2013dynamic}. Learning only from short-range temporal interactions (e.g 30-40 seconds as typically done when cropping or defining dynamic FC windows) ignores the evolution of the signal and the sequential switching through cognitive states and is more susceptible to psychological noise.

Apart from capturing the true underlying dynamics of the data, a key consideration when developing machine learning models is their utility and adoption by the community. Today, for a researcher interested in investigating a certain phenotype or a clinical outcome using machine learning, the main go-to remains FC analysis using linear models or \textit{shallow} kernel based methods. The main reasons for this are i)\textbf{Simplicity}: easy to implement and interpret without much engineering and hyperparameter tuning, ii)\textbf{Sample-efficiency}; possible to train with small sample sizes as is often the case in clinical datasets. iii) \textbf{Computational-efficiency}; do not require special hardware. iv)\textbf{Performance}; most importantly is that they achieve the desired objective. Several recent studies have reported competitive performance of linear models or shallow non-linear models against DL in phenotype prediction from neuroimaging data \cite{schulz2020different, he2020deep}.

To bridge this gap and to address the dynamic limitation in several DL architectures, we propose fMRI-S4; a powerful deep learning model that leverages 1D-CNNs and state-space models to learn short- and long-range saptio-temporal features from rs-fMRI data. fMRI-S4 is open-source, data-efficient, easy to train, and can outperform existing methods on phenotype prediction without requiring special hyperparameter tuning for each task/dataset/ROI-template. We validate our work on three multi-site datasets encompassing three different targets. Namely, the \textit{\textbf{UkBiobnak}\cite{ukbiobank}} for sex classification, \textit{\textbf{ABIDE}\cite{abide}} for autism ASD diagnosis, and \textit{\textbf{Rest-Meta-MDD}}\cite{restmdd} for MDD diagnosis.

\section{Methodology}

\subsection{Preliminaries}
Phenotype prediction from rs-fMRI data can be formulated as a multivariate timeseries classification problem. Let $X$ $\in$ $R^{N \times T}$ represent a resting-state functional scan where $N$ denotes the number of brain ROIs as extracted using a pre-defined template (\textit{The spatial dimension}), and $T$ represent the number of timepoints sampled for the duration of the scan (\textit{The temporal dimension}). Given a labelled multi-site dataset $D = \{X_{i},Y_{i}\}_{i=1}^{S}$, where $Y$ is a non-imaging phenotype or a disorder diagnosis, the objective is to learn a parameterized function $f_{\theta}$ $\colon$ $X$ $\mapsto$ $Y$.This is under the practical limitations of small sample size $S$(typically in the order of hundreds in rs-fMRI multi-site datasets), and that scan duration $T$ and temporal resolution $Tr$ are variable within the dataset since it is collected from different scanning sites. The main challenge then becomes how to design a generalizable $f_{\theta}$ that can capture the underlying causal variables in the data necessary to predict the target. 

\subsection{Learning short-range dependencies with 1D Convolutions}
The resting-state signal is characterized by low-frequency oscillations (0.01-0.1 Hz). To extract informative features from such signal, a model has to learn the short-range dynamic dependencies that characterize the neural activations responsible for generating the oscillations. While in essence, a feature extraction layer that scans the entire global signal (e.g. Transformers, Fully connected layers, RNNs) can learn such local dependencies, leveraging the inductive bias of convolutions significantly improve sample- and parameter- efficiency of the model. Convolutional layers excel at learning local patterns using a translation invariant sliding window and have been used in conjugation with \textit{global} layers to alleviate the memory bottleneck and enhance locality \cite{li2019enhancing, shi2015convolutional}.

We utilize a convolutional encoder as first stage in the model to learn local dependencies, improve the signal to noise ratio and to mix the features in the spatial dimension.  Our encoder consists of $K_{conv}$ blocks, where each blocks consists of a 1D convolutional layer with kernel size $k$ and hidden dimension $d_model$, followed by batch normalization, and then \textit{relu} activation. 1D CNNs treat the $N$ ROIs as input channels and thus the temporal kernel is fully connected across the spatial dimension. This is a very useful property since the ordering of the ROIs is arbitrary and the model is implicitly free to learn any spatial dependencies necessary for the objective regardless of the parcellation template applied during pre-processing. 

\subsection{Learning Long-range dependencies with State Space models}
1D convolutions are a  powerful tool to model spatio-temporal dependencies. Yet, they are limited by their receptive field in the temporal dimension. While this can be improved by adding more layers, increasing the kernel size or using dilation \cite{oord2016wavenet}, the output features at each layer remain constrained by the receptive field of the respective convolutional filter size. Given that 1DCNN models are typically small (1-3 layers)\cite{wang2017time,el2019simple}, the output features only represent local dependencies. Further, since fMRI datasets are usually collected at multiple locations using different scanners, the temporal resolution of the data may vary across the dataset, which deems finding an optimal kernel size and dilation factors a challenging engineering task. 
A state space model (SSM) is a linear representation of dynamical systems in continuous or discrete form. The objective of SSMs is to compute the optimal estimate of the hidden state given the observed data\cite{williams2007linear}. SSMs have been widely in modelling fMRI signal, and its applications includes decoding mental state representation\cite{hutchinson2009modeling, eavani2013unsupervised, janoos2011spatio}, inferring effective connectivity\cite{tu2019state}, generative models for classifcation\cite{suk2016state}. Inspired by the recent developments in SSMs, specifically the \textbf{S4} model\cite{s4}, which utilizes state-space models as sequence-to-sequence trainable layers and excel at long-range tasks, we propose to integrate S4 layers in our fMRI classifier to capture the global dependencies in the signal for single-subject prediction of non-imaging phenotypes. 
SSM in continuous time-space maps a 1-D input signal $u(t)$ to an M-D latent state $z(t)$ before projecting to a 1-D output signal $y(t)$ as follows:

\begin{align}
\begin{split}\label{eq:1}
    z^{'}(t) = Az(t) + Bu(t)
\end{split}\\
\begin{split}\label{eq:2}
    y(t) = Cz(t) + Du(t)
\end{split}
\end{align}
Where $A \in R^{M \times M}$ is the  state-transition matrix, $B \in R^{M \times 1}$, $C \in R^{1 \times M}$ and $D \in R^{1 \times 1}$ are projection matrices. 
To operate on discrete-time sequences sampled with a step size of $\Delta$, the SSM can be discretized using the bilinear method \cite{bilinear} as follows:
\begin{align}
\begin{split}\label{eq}
    z_{k} = \bar{A}z_{k-1} + \bar{B}u_{k}  \qquad    y_{k} = \bar{C}z_{k} + \bar{D}u_{k}
\end{split}
\end{align}
\begin{align}
\begin{split}\label{eq:4}
    \bar{A} = (I- \Delta/2 \cdot A)^{-1} (I + \Delta/2 \cdot A ) \qquad  \bar{B} = (I- \Delta/2 \cdot A)^{-1} \Delta B ) \qquad \bar{C} = C 
\end{split}
\end{align}
Given an initial state $z_{k} = 0$ and omitting $D$ (as it can be represented as a skip connection in the model), unrolling \ref{eq} yields:
\begin{align}
\begin{split}\label{eq:5}
   y_{k} = \bar{C}\bar{A}^k\bar{B}u_{0} + \bar{C}\bar{A}^{k-1}\bar{B}u_{1} + ... + \bar{C}\bar{B}u_{k}\\
   y = \bar{K} * u  \qquad \bar{K} = (\bar{C}\bar{A}^i\bar{B})_{i \in [L] }
\end{split}
\end{align}
 
\begin{figure}[!ht]
\includegraphics[width=\textwidth, scale=0.5]{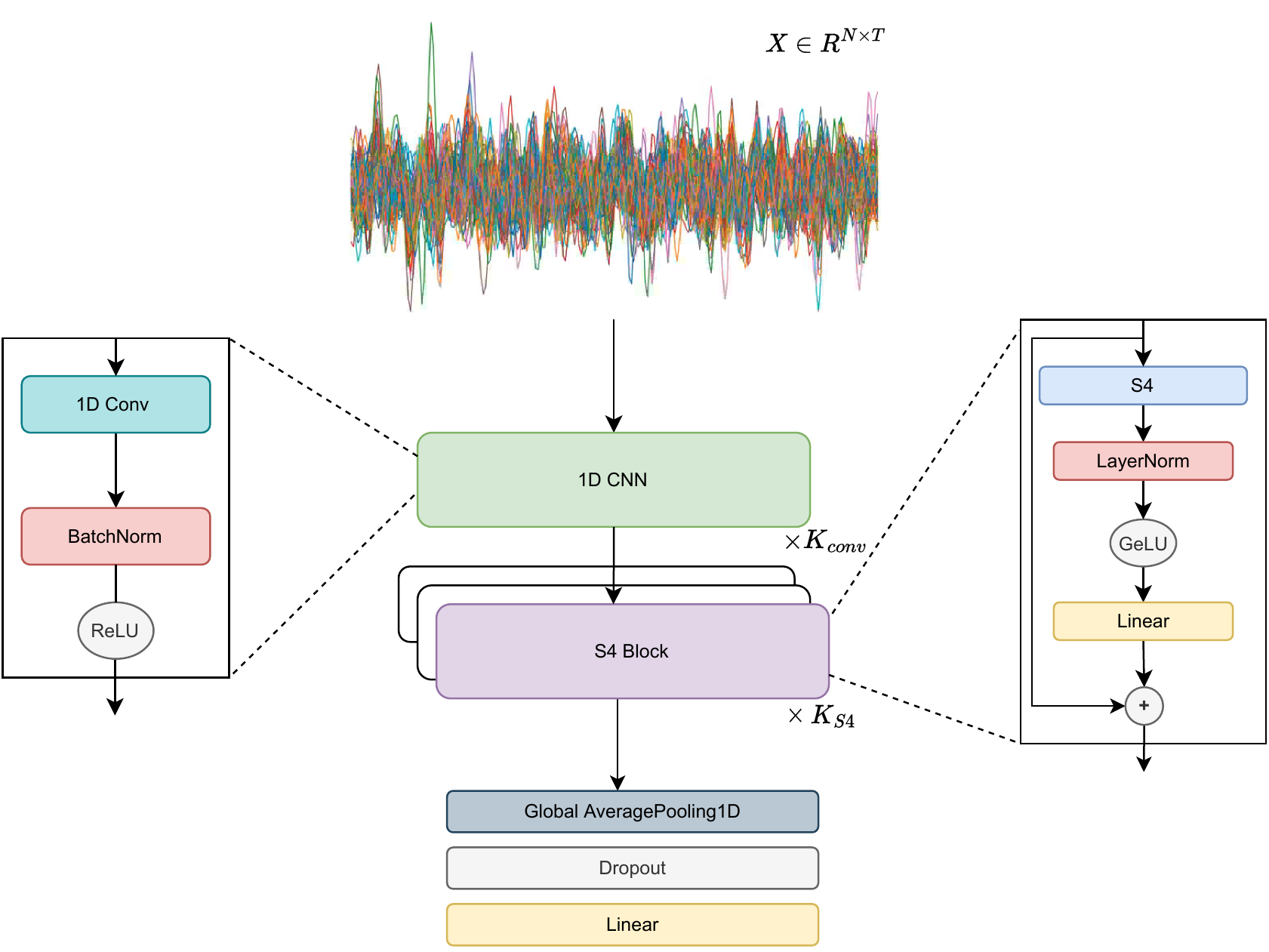}
\centering
\caption{An overview of the fMRI-S4 Model. The 1D CNN Blocks learn short-range temporal dependencies using a small kernel size and learn spatial dependencies across the ROIs. The output features are then fed to a cascade of S4 Blocks to learn both short- and long-range temporal dependencies.}\label{mainfig}
\end{figure}

The operator $\bar{K}$ can thus be interpreted as a convolution filter and the state space model can be trained as a sequence-to-sequence layer via learning parameters $A$,$B$,$C$ and $\Delta$ with gradient descent. Training $\bar{K}$ efficiently requires several computational tricks. The S4 paper proposes the parameterization of $A$ as a diagonal plus low-rank (DPLR) matrix. This parameterization has two key properties. First, this is a structured representation
that allows faster computation using the Cauchy-kernel algorithm \cite{cauchy2} to compute the convolution kernel $K$ very
quickly. Second, this parameterization includes certain special matrices called HiPPO matrices \cite{hippo}, which
theoretically and empirically allow the SSM to capture long-range dependencies better via memorization. For in-depth details of the model we refer the readers to \cite{annotateds4}. SSMs defines a map from $R^L$ $\mapsto$ $R^L$, i.e. a 1-D sequence map. To handle multi-dimensional inputs/features  $R^{L \times H}$, the S4 layer simply defines $H$ independent copies of itself at  and after applying a non-linear activation function and layer normalization,  the $H$ feature maps are mixed with a position-wise linear layer. This defines a single S4 block. In our model, we stack $K_{s4}$ blocks on top of the convolutional layers, followed by a global average pooling layer across the temporal dimension, a dropout layer and a Linear layer with output dimension equal to the number of classes. Fig. \ref{mainfig} demonstrates the overall architecture of the model.

\subsection{fMRI-S4: Towards a fixed baseline}\label{section}

A key aspect when designing a classification model for fMRI is versatility. Ideally, we would like to use fixed architecture and fixed training parameters for any dataset/any target and obtain competitive results without the need of exhaustive hyperparameter search and model enigeering. This would improve the utility and accessibility of the model for practitioners with different technical backgrounds, and facilitate models benchmarking. The proposed fMRI-S4 model constitutes several desirable properties that makes it feasible to find such optimal architecture. The sliding window approach in 1D convolutions enable feature extraction independent of the length of the time course. Further, it's permutation invariant in the spatial dimension, i.e. it  can map any number of ROIs with any arbitrary ordering into a fixed dimension. An S4 layer learn a global representation of the data, and can thus eliminate the need to tailor the number of layers in the model to cover the desired receptive field. Moreover, S4 learn an adaptive discritization step size $\Delta$, which further improve the flexibility of the model with respect to the variable temporal resolution of the datasets. This setup in turn can improve the odds of finding a set of optimal parameters for the model and training that can achieve competitive results invariant to the task and the dataset. To this end, we conducted Bayesian hyperparameter tuning using the weights\&biases platform \cite{wandb} on independent validation sets on the three datasets presented in section \ref{datasets} to find the optimal configuration of the model. The top performing configurations converged to a highly overlapping set and based on this we report our default configuration of the model which we use for all the experiments in this work and recommend as a baseline. For the model architecture we use $d_{model}$ = 256, $K_{conv}$ = 1 , $K_{S4}$ = 2 , $d_{state}$ = 256. The rest of the parameters are fixed as in the original S4 model. 
We trained the models using the cross-entropy loss optimized using AdamW Optimizer with a learning rate of 1e-4 and a weight decay of 1e-5. The training is run for 100 epochs with early stop with patience=10 conditioned on the best accuracy of an inner-validation set. Training the model takes $\approx$ 15-20 minutes on a Nividia 16-GB P100 GPU. This configuration constitutes 1.3 M trainable parameters. 

\section{Experiments \& Results}

\subsection{Datasets}\label{datasets}
To evaluate the performance of the the proposed model, we utilized three multi-site rs-fMRI datasets each addressing a different objective. 

1) \textit{\textbf{Rest-Meta-MDD}} \cite{restmdd} is currently the largest open-source rs-fMRI database for studying major depressive disorder including clinically diagnosed patients and healthy controls from 25 cohorts in China. In this work, we use a sample of  $N$ $=$ 1453 (628 HC/825 MDD) which survived the quality check to evaluate the model performance on the task of MDD diagnosis. See \cite{restmdd} for the exact pre-processing pipeline. We adopted the Harvard Oxford (HO) \cite{ho} atlas to segment the brain into 118 cortical and subcortical ROIs. \newline
2) \textit{\textbf{ABIDE I+II}} \cite{abide} contains a collection of rs-fMRI brain images aggregated across 29 institutions. It includes data from participants with autism spectrum disorders and typically developing participants (TD). In this study, we used a subset of the dataset with $N$ $=$ 1207 (558 TD/649 ASD) to evaluate the model performance on the tasks of ASD diagnosis. We utilized the C-PAC pre-processing pipeline to pre-process the data. Similarly, we adopted the HO atlas to segment the brain.\newline
3) \textit{\textbf{UkBioBank}} \cite{ukbiobank} is a large-scale population database, containing in-depth genetic and health information  from  half  a  million  UK  participants.  In  this  work  we  use  a  randomly sampled subset (N =5500 ( 2750 M/ 2750 F ) to evaluate the model performance on the sex classification task. The ROIs were extracted using the Automated Anatomical Labeling (AAL)  \cite{aal} atlas.

\subsection{Clinical Results}

We evaluated the performance of the fMRI-S4 model against following baselines: \textbf{SVM},  \textbf{BrainNetCNN}\cite{kawahara2017brainnetcnn}, \textbf{1D-CNN}\cite{el2019simple}, \textbf{ST-GCN}\cite{gadgil2020spatio} and \textbf{DAST-GCN}\cite{el2021dynamic}. The input representation to the  \textbf{SVM} and \textbf{BrainNetCNN} are the static Person correlation matrices of the ROIs timecourses, while the rest of the methods operate directly on the ROIs timecourses. For a fair evaluation, we conduct a hyperparameter search using  weights$\&$biases \cite{wandb} for the baselines on a independent validation set to select the best configuration for each task. Next, we conducted the experiments using a repeated 5-fold cross validation scheme on the two clinical tasks using the selected parameters for the baselines and the fixed configuration presented in Section \ref{section} for fMRI-S4.  We report the results in Table \ref{table1}. For both tasks, fMRI-S4 outperform the best performing baseline (the 1D-CNN) by 1.6 accuracy points (2.5 \% relative) and 2.4 accuracy points (3.4 \% relative) on the \textit{\textbf{Rest-Meta-MDD}} and \textit{\textbf{ABIDE I+II}} datasets, respectively. To demonstrate the efficacy of combining both 1D convolutions and state space models, we conduct a simple ablation study with two models, i)fMRI-S4$_{K_{S4}=0}$; where the S4 layers are removed and replaced by 2 convolutional layers. ii) fMRI-S4$_{K_{conv}=0}$; where no convolution layers are used, and spatio-temporal feature extraction is done using S4 layers only. In both cases, the accuracy of the model drops by 2-4 \% for both datasets. 


\begin{table}
\caption{5-fold test metrics for fMRI-S4, ablated version of fMRI-S4 and baseline models on the two clinical datasets.}\label{table1}
\centering
\begin{tabular}{|l|lll|lll|}

\hline
\multicolumn{1}{|c|}{\multirow{2}{*}{Model}} & \multicolumn{3}{c|}{\textit{\textbf{Rest-Meta-MDD}}}                                                   & \multicolumn{3}{c|}{\textit{\textbf{ABIDE I+II}}}                                                       \\ \cline{2-7} 
\multicolumn{1}{|c|}{}                       & \multicolumn{1}{l|}{Acc.(\%)}         & \multicolumn{1}{l|}{Sens.(\%)}        & Spec.(\%)        & \multicolumn{1}{l|}{Acc.(\%)}          & \multicolumn{1}{l|}{Sens.(\%)}        & Spec.(\%)        \\ \hline
SVM                                      & \multicolumn{1}{l|}{61.9$\pm$3}   & \multicolumn{1}{l|}{73.0$\pm$9}   & 50.9$\pm$8   & \multicolumn{1}{l|}{69.5$\pm$3}  & \multicolumn{1}{l|}{75.9$\pm$5} & 63.0$\pm$5 \\
BrainNetCNN\cite{kawahara2017brainnetcnn}                                  & \multicolumn{1}{l|}{58.4$\pm$5}   & \multicolumn{1}{l|}{50.1$\pm$11}  & 56.8$\pm$8 & \multicolumn{1}{l|}{66.6$\pm$3}  & \multicolumn{1}{l|}{72.1$\pm$4} & 61.2$\pm$5 \\
ST-GCN \cite{gadgil2020spatio}         & \multicolumn{1}{l|}{58.2$\pm$4}   & \multicolumn{1}{l|}{48.6$\pm$9} & 67.8$\pm$7 & \multicolumn{1}{l|}{65.3$\pm$2}  & \multicolumn{1}{l|}{67.2$\pm$3} & 63.4$\pm$4 \\
DAST-GCN\cite{el2021dynamic}         & \multicolumn{1}{l|}{60.7$\pm$4} & \multicolumn{1}{l|}{44.6$\pm$7} & 75.8$\pm$7 & \multicolumn{1}{l|}{67.8$\pm$2}  & \multicolumn{1}{l|}{70.8$\pm$3} & 64.9$\pm$3 \\
1D-CNN\cite{el2019simple}             & \multicolumn{1}{l|}{63.8$\pm$2} & \multicolumn{1}{l|}{65.7$\pm$3} & 61.9$\pm$3 & \multicolumn{1}{l|}{70.6$\pm$2}   & \multicolumn{1}{l|}{72.7$\pm$3} & 68.5$\pm$3 \\ \hline
fMRI-S4$_{K_{S4}=0}$           & \multicolumn{1}{l|}{63.7$\pm$3} & \multicolumn{1}{l|}{68.5$\pm$6} & 59.7$\pm$6 & \multicolumn{1}{l|}{71.7$\pm$2}  & \multicolumn{1}{l|}{75.7$\pm$5} & 67.8$\pm$3 \\
fMRI-S4$_{K_{conv}=0}$           & \multicolumn{1}{l|}{62.9$\pm$2}  & \multicolumn{1}{l|}{67.5$\pm$5} & 60.1$\pm$5 & \multicolumn{1}{l|}{70.4$\pm$3} & \multicolumn{1}{l|}{74.5$\pm$3} & 66.1$\pm$1 \\
fMRI-S4            & \multicolumn{1}{l|}{\textbf{65.4$\pm$3}} & \multicolumn{1}{l|}{66.9$\pm$4} & 63.9$\pm$4 & \multicolumn{1}{l|}{\textbf{73.0$\pm$3}}  & \multicolumn{1}{l|}{75.2$\pm$4} & 70.8$\pm$2 \\ \hline
\end{tabular}

\end{table}

\section{Evaluating sample-efficiency with the UkBioBnak}
\begin{figure}[t]
\includegraphics[width=\textwidth]{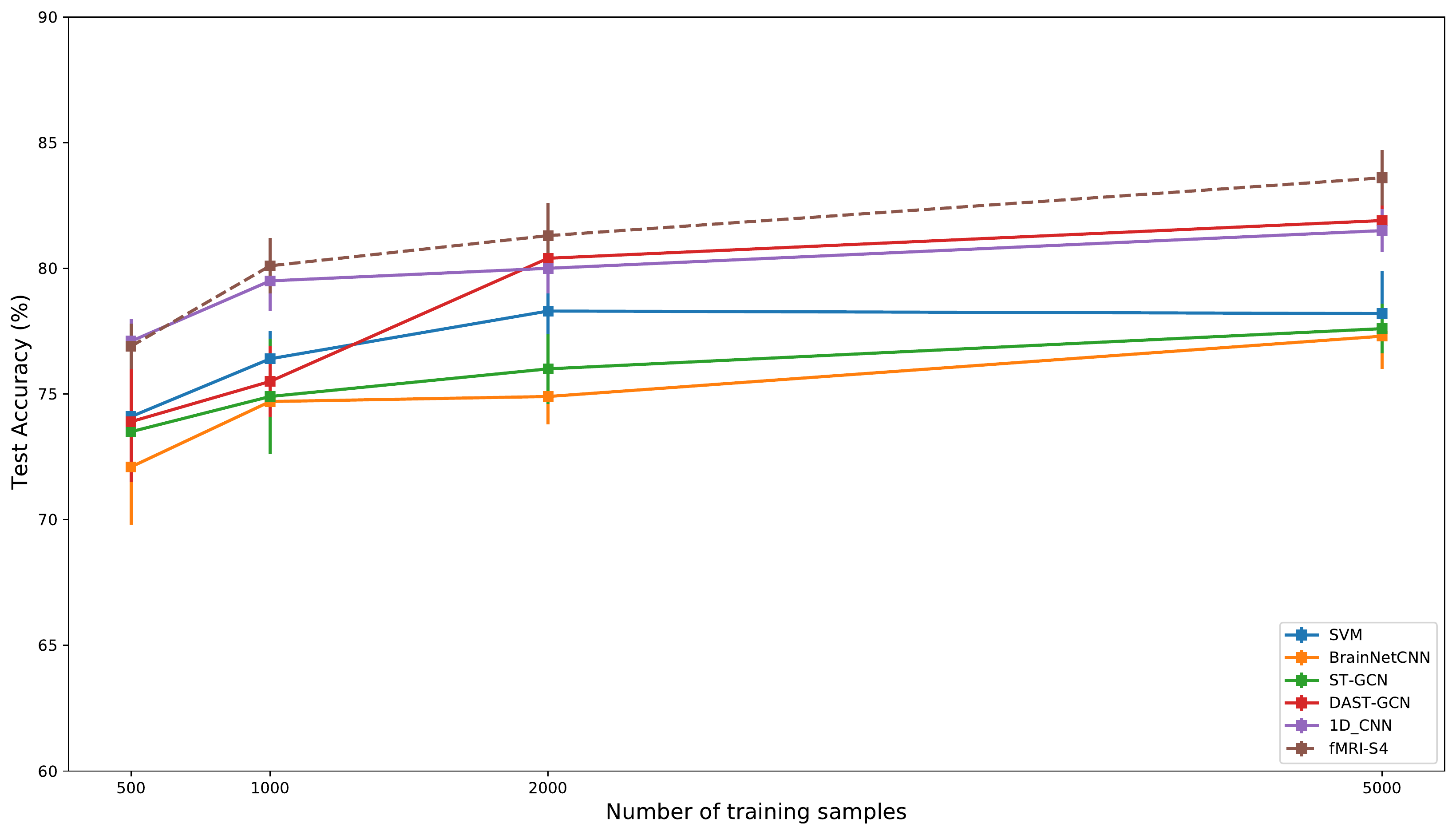}
\centering
\caption{Scaling performance of the fMRI-S4 against existing baselines on the task of sex classification on the UkbioBank dataset. The error bars represent results for 3 different random seeds for initialisation and inner validation split. The test set is fixed for all the experiments.}\label{fig2}
\end{figure}

To compare the sample efficiency of fMRI-S4 against existing baseline, we conduct a training sample scaling experiment on the UkbioBank dataset for the task of sex classification. Namely, we train the models using N=[500,1000,2000,5000] class-balanced samples, and evaluate the performance of the trained models on a fixed test set with N=500 (250 M/ 250F). The results presented on Fig. \ref{fig2} highlight that fMRI-S4 performs competitively at the smallest sample sizes (N=500) and continue to scale favourably against the baselines. Another observation from the results is the scaling superiority of the dynamic models (1D-CNNs, fMRI-S4, DAST-GCN) over static models with exception of the ST-GCN model. This behaviour suggests that the evolution of the dynamic signal contain discriminate information that can be better exploited by the deep learning models at larger training samples.

\section{Discussion}
In this work, we present fMRI-S4; a deep learning model that leverages 1D convolutions and state-space models for learning short- and long-range dependencies necessary to capture the underlying dynamic evolution of the brain activity at rest. We show fMRI-S4 improves the diagnosis of MDD, ASD and sex classification from rs-fMRI data against existing methods using a fixed architecture for all three tasks. We hope that this work can improve the adoption of dynamic DL-based models in fMRI analysis and motivate the development of generalizable methods. In our future work, we will investigate dynamic feature perturbation to explain the predictions of fMRI-S4 in effort to obtain potential biomarkers for psychiatric disorders.

\bibliographystyle{splncs04}
\bibliography{main.bib}
\end{document}